\documentclass{bioinfo}
\copyrightyear{2005}
\pubyear{2005}

\begin{document}
\firstpage{1}

\title[short Title]{A study of structural properties on profile HMMs}
\author[Sample \textit{et~al}]{Juliana S Bernardes\,$^{\rm a}$\footnote{to whom correspondence should be addressed}, Alberto M. R. D\'avila\,$^{\rm b}$, V\'itor Santos Costa$^{\rm c}$,  Gerson Zaverucha$^{\rm a}$}

\address{$^{\rm a}$COPPE Engenharia de Sistemas e Computa\c c\~ao, UFRJ, Rio de Janeiro,
 $^{\rm b}$Instituto Oswaldo Cruz, Fiocruz, Rio de Janeiro, Brazil,
 $^{\rm c}$CRACS and DCC-FCUP, Universidade do Porto, Portugal}
\maketitle

\begin{abstract}

\section{Motivation:}
Profile hidden Markov Models (pHMMs) are a popular and very useful tool in the detection of the remote homologue protein families. Unfortunately, their performance is not always satisfactory when proteins are in the ``twilight zone''. We present HMMER-STRUCT, a model construction algorithm and tool that tries to improve pHMM performance by using structural information while training pHMMs. As a first step, HMMER-STRUCT constructs a set of pHMMs. Each pHMM is constructed by weighting each residue in an aligned protein according to a specific structural property of the residue. Properties used were primary, secondary and tertiary structures, accessibility and packing. HMMER-STRUCT then prioritizes the results by voting.

\section{Results:}
We used the SCOP database to perform our experiments.  Throughout, we apply leave-one-family-out cross-validation over protein superfamilies. First, we used the MAMMOTH-mult structural aligner to align the training set proteins. Then, we performed two sets of experiments. In a first experiment, we compared structure weighted models against standard pHMMs and against each other. In a second experiment, we compared the voting model against individual pHMMs.  We compare method performance through ROC curves and through Precision/Recall curves, and assess significance through the paired two tailed t-test. Our results show significant performance improvements of all structurally weighted models over default HMMER, and a significant improvement in sensitivity of the combined models over both the original model and the structurally weighted models.

\section{Availability:}
The HMMER-STRUCT tool has been implemented as Perl scripts and as C source code. The structure weighting procedure is available as a patch to the HMMER program. All the test sets, train sets, programs and scripts used in this study are available in
\href{http://wiki.biowebdb.org/index.php/Hmmer-struct}{http://wiki.biowebdb.org/index.php/Hmmer-struct}.

\section{Contact:} \href{julianab@cos.ufrj.br}{julianab@cos.ufrj.br}
\end{abstract}

\section{Introduction}

One of the major tasks in computational molecular biology is to aid large-scale protein annotation and biological knowledge discovery. Functional characterization of unknown-function proteins is often inferred through sequence similarity search methods, such as BLAST \citep{BLAST70} and FASTA \citep{FASTA85}. However, when the evolutionary relationship among proteins is distant, methods based on profile hidden Markov models (pHMMs) \citep{PHMMIntroEddy96,PHMMIntroKrogh94} are known to outperform  methods based on sequence similarity search \citep{Gough01,Park98}.

Profile Hidden Markov Models are probabilistic models that are often used to represent groups of homolog sequences. These models have been a key tool in protein annotation, and are highly effective for scoring similar sequences. Unfortunately, the performance of pHMMs degrades for sequences in the twilight zone, that is, for homologue sequences with low identity (below 30\%). This limitation has motivated a number of different approaches to increase pHMM performance. Proposals include new scoring functions, new null models \citep{NewNullModel05} and prior probability \citep{Dirichlet93}. Researchers have also combined other information with pHMMs: T-HMM \citep{THMM04} uses phylogenetic information;  HMM-STR~\citep{HMMSTR00},  combines pHMMs and support vector machines~\citep{SVM99}.

The observation that homologue proteins tend to preserve structure suggests that structural information should be extremely relevant in detecting homologues. In fact, it has been shown that pHMMs trained with multiple sequence alignments based on protein’s structural alignment can have better performance than  pHMMs based on state-of-the art aligners that apply primary sequence information only, when remote homology detections are assessed~\citep{Bernardes07}. In this vein, researchers have  proposed special alphabets to represent structural elements in pHMMs \citep{SASearch_04}, or modifying pHMM structure to add protein three-dimensional information \citep{Alexandrov04}. Although such methods are more powerful than pHMMs, arguably they are  computationally more expensive both in training and in classification, and to the best of our knowledge have not become widely used.

We present a novel method to apply structural information in protein classification. In contrast to the previous approaches, our method relies on pHMMs. Our main contribution is a residue weighting-algorithm that incorporates protein structural information into pHMMs. Further, we apply different structural properties to train a library of 5 pHMMs from a homologue protein set. The properties we consider are primary, secondary, and tertiary structure, also used in previous methods. We also apply two properties that, to the best of our knowledge, have not been used in this task before, but that are often important in this domain: solvent accessibility and residue packing. The classification of a unknown-function protein
is then obtained by combining the classification from the library of pHMMs. The main advantage of our method is that structural information is only used to train the pHMMs. Notice that scoring is still performed using sequence data, as opposed to \citep{Alexandrov04}. Our method was implemented by extending the HMMER package and experimental evaluation using the SCOP database showed significant improvement over HMMER.

\begin{methods}
\section{Methods}

In our experiments, a protein homologue set is aligned by utilizing the MAMMOTH-mult~\citep{Mammoth05} aligner. MAMMOTH produces two outputs: one represents the multiple sequence alignment (based on spatial coordinates’ similarity) and the other the structural alignment. These outputs are used to build weight matrices $M^{'}_{s}$, which represent residue structural weights for each protein. These  matrices were used on pHMM training stage. The section \ref{ss:sq} will give more details on the building of $M^{'}_{s}$. Basically, our approach builds five pHMMs. The simplest pHMM is built from MAMMOTH's multiple sequence alignment, by keeping the default sequence-weighting algorithm of HMMER. This model is called \textit{pHMM1D}. In order to aid to build the remaining pHMMs our approach generates $M^{'}_{s}$. The matrices used in building \textit{pHMM2D},  \textit{pHMMAcc}  and  \textit{pHMMOi}, incorporate secondary structure, residue solvent accessibility, and residue packing information, respectively. In order to build these matrices were used both MAMMOTH sequence alignment plus structural properties obtained using the \textit{joy} package \citep{Joy98} to collect these information from PDB coordinates \citep{Pdb00}. Last, we used MAMMOTH's multiple sequence alignment plus structural alignment to build a matrix based on homologue core structures~\citep{Matsuo99}. That matrix was used to build \textit{pHMM3D}. Figure \ref{fig:hmmer_struct} shows the proposed method.

\begin{figure}[!h]
\centering
 \includegraphics[scale=0.5]{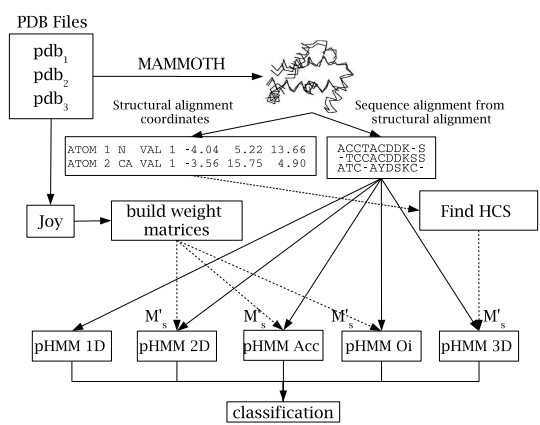}
\caption{First, a homologue protein set is aligned by using MAMMOTH-mult aligner. The aligner produces a multiple sequence alignment and a structural alignment. The multiple sequence alignment is used to build a conventional pHMM using HMMER package, \emph{pHMM1D}. Aligned sequences are fed to the \textit{joy} tool. The \textit{joy} output is used to construct weight matrices, which are then used to build secondary \emph{pHMM2D}, accessibility \emph{pHMMAcc}, and packing \emph{pHMMOi} models.  Finally, the structural alignment is used to find the homologue core structure, which is then used to construct \emph{pHMM3D}.}\label{fig:hmmer_struct}
\end{figure}

\subsection{Profile HMMs}

Profile HMMs represent conserved regions in sequences as sequences of \emph{match} (M) states. Inserted material is represented as \emph{insert} states (I), and deleted regions as \emph{delete} states (D).  The parameters of pHMMs are probabilities of two events: a \emph{transition probability} from a state to another state, and a probability that a specific state will emit a specific residue (say, a specific amino-acid when comparing proteins), called \emph{emission probability}. Obviously, only match and insert states generate characters and have an emission probability distribution; delete states are quiet. In the case of proteins, emission distributions have 20 entries, one per amino-acid.

Possible transitions define the structure of the pHMM. Systems such as SAM \citep{SAM96} allow transitions between all types of states, totaling 3 transitions per state, hence 9 per node. On the other hand, the HMMER system relies on the Plan7 model \citep{ProfileEddy98}, which disallows $I \rightarrow D$ and $D \rightarrow I$ transitions.

Emission probabilities are calculated by the equation \ref{eq:prob_em}, where $c_{j}(\sigma)$ is the observed frequency of residue $\sigma$ in $j$ column of the alignment, and $\alpha(\sigma)$ represents the pseudo counts of residue $\sigma$, which are obtained from Dirichlet mixtures, as seen in \citep{Dirichlet93}.

\begin{equation}
\label{eq:prob_em}
\\e_{j}(\sigma) = \dfrac{c_{j}(\sigma) + \alpha(\sigma)}{\sum_{k} c_{j}(\sigma_{k}) + \alpha(\sigma_{k})}
\end{equation}

In the same way, the transition probability can be found through the equation \ref{eq:prob_trans}, where $c_{kl}$ is the observed frequency of transitions between state $k$ and state $l$, where $k,l \in \{M, I, D\}$, and $\alpha_{kl}$ represents the pseudo counts of transition between $k$ and $l$.

\begin{equation}
\label{eq:prob_trans}
\\t_{kl} = \dfrac{c_{kl} + \alpha_{kl}}{\sum_{l} c_{kl} + \alpha_{kl}}
\end{equation}

\subsection{Sequence Weighting}
\label{ss:sq}

One problem in representing families of sequences is that often sets of very similar sequences may be over-represented in the training sequences, introducing bias. Therefore, \emph{sequence-weighting} methods were introduced to compensate for over-representation among multiply aligned sequences. In general, very similar sequence receives lower weights and divergent sequence higher weight. Sequence weighting was applied to the construction of position-specifics score matrix (PSSM) \citep{Gribskov87}, and is fundamental to the performance of profile HMMs. In the latter case, the default sequence weighting method used by HMMER package is a high quality algorithm based on phylogenetic trees~\citep{SeqWeiGSC94}.

Let $A$ be an generic alignment used to train a pHMM. Suppose, $A$ with $N$ sequences and length $L$. Then, we can represent $A$ alignment weights as a matrix $W$, such that $w_{ij}$ represents the weight of an amino-acid of protein $I$ in the $j^{th}$ alignment position, as shown in the equation \ref{eq:matriz_weight}. Basically, a sequence-weighting method for pHMMs attributes equal weights to all residues in the protein, that is, $w_{ij} = w_{ik}$ for $\forall j,k \leq L$.
\\

\begin{equation}
\mathbf{W=}
\left(\begin{array}{ccc}
w_{11} & \ldots & w_{1L} \\
\vdots & \vdots & \vdots \\
w_{N1} & \ldots & w_{NL} \\
\end{array}\right)
\label{eq:matriz_weight}
\end{equation}
\\

In the spirit of PSSMs, we propose to reinforce residues that correspond to preserved regions in the protein. Our motivation is that when homologue proteins are structurally aligned, spatial overlapping of an atom set occurs. This set is called the \emph{invariant core} or \emph{core structure}, and can be used to characterize homologue proteins. We argue that the residues in the core structure should carry more weight rather than the residues outside the core. Thus, we propose sequence-weighting method  that gives different weight to each residue in the same protein, based on structural relevance. We will represent such ``structural'' weights by a matrix $M_{s}$, where each residue of the same protein has a different weight.
\\

\begin{equation}
\mathbf{M_{s}=}
\left(\begin{array}{ccc}
m_{11} & \ldots & m_{1L} \\
\vdots & \vdots & \vdots \\
m_{N1} & \ldots & w_{NL} \\
\end{array}\right)
\label{eq:matriz_struct}
\end{equation}
\\

As mentioned before, the default sequence-weighting method used by HMMER package is a high quality algorithm. Therefore, we combine both the default HMMER's $M$ matrix in (\ref{eq:matriz_weight}) and the of $M_{s}$ structural matrix in (\ref{eq:matriz_struct}), as shown in (\ref{eq:matriz_struct_hmmer}).
\\

\begin{equation}
\mathbf{M^{'}_{s}=MM_{s}^{T}}=
\left(\begin{array}{ccc}
w_{11}m_{11} & \ldots & w_{1L}m_{1L} \\
\vdots & \vdots & \vdots \\
w_{N1}m_{N1} & \ldots & w_{NL}m_{NL} \\
\end{array}\right)
\label{eq:matriz_struct_hmmer}
\end{equation}
\\

However, introducing weights affect the computation of the observed frequencies. More precisely, the observed frequency $c_{j}(\sigma)$ shown in \ref{eq:prob_em} is now found through the equation \ref{eq:count_em}, where $s_{ij} =  w_{ij}m_{ij}$ is structural weight of residue $\sigma$, according to $M^{'}_{s}$ matrix.

\begin{equation}
\label{eq:count_em}
\\c_{j}(\sigma) = \sum_{i}^{N} f(\sigma)\therefore
f(\sigma) = \left\{ \begin{array}{cc}s_{ij}, & \textrm{if } \sigma\textrm{ is the amino-acid} \textrm{ in position ij} \\
\\ 0, & \textrm{otherwise}
\end{array} \right.
\end{equation}
\\

In the same way, we apply the equations \ref{eq:count_trans} to determine $c_{kl}$ shown in \ref{eq:prob_trans}. If the $k$ and $l$ states are either M or I states, $c_{kl}$ can be calculated through the arithmetic mean of $m_{ik}$ and $m_{il}$. If at least one state is a D state, $c_{kl}$ is either $m_{ik}$, if $l \in \{D\}$, or $m_{il}$, if $k \in \{D\}$. Last, if both are D states, $c_{kl}$ is 1.

\begin{equation}
\label{eq:count_trans}
\\c_{kl} = \sum_{i}^{N} f_{kl}\therefore
\\f_{kl} = \left\{ \begin{array}{cc} \dfrac{s_{ik} + s_{il}}{2}, & \textrm{se k,l} \in {\{M,I\}} \\
\\s_{ik}, & \textrm{se l} \in {\{D\}} \ e \ \textrm{k} \notin {\{D\}}\\
\\s_{il}, & \textrm{se k} \in {\{D\}} \ e \ \textrm{l} \notin {\{D\}}\\
\\ 1, & \textrm{se k,l} \in {\{D\}}
\end{array} \right.
\end{equation}
\\

\subsection{The $M_{s}$ structural weight matrices}

As explained above, our algorithm considers a number of different sources of structural information.
Next, we approach how this information was obtained and used to built $M_{s}$ matrix.

\subsubsection{Secondary structural elements}

Secondary structure is often conserved among homologue proteins. Indeed, \emph{motifs} \citep{Branden91}, consensus sequences in homologue proteins, usually include a combination of well conserved secondary structure elements \citep{Chakrabarti04}.

In order to build a $M_{s}$ matrix based on secondary structure elements we need to identify secondary structure elements in the original sequences. This is possible because we assume we have full structural data for the \emph{training} sequences. In this work, we chose to utilize the SSTRUCT program, part of the widely used \texttt{joy} package~\citep{Joy98}, to extract secondary elements from the PDB files.  SSTRUCT output is a character sequence, such that the characters \{L=loop, H=helix, C=sheet\} match a secondary structure element against a residue, as shown in figure \ref{fig:sq_mat_struc_sec}.  Following Deane's work on the relative frequency of conserved regions \citep{Deane03}, we mapped each SSTRUCT element as follows: $L \rightarrow 1$, $H \rightarrow 2$, and $ C \rightarrow 4$. Our mapping thus favours conservation in sheets, and gives default weight to loops. Although the active site of proteins can be found in loops, these regions often contain \textit{indel} segments. Figure~\ref{fig:sq_mat_struc_sec} shows an example of structural weight attributions for proteins in a partial alignment.
\\

\begin{figure}[!h]
\centering
\includegraphics[scale=0.5]{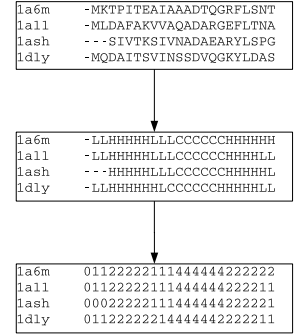}
\caption{Representation of the secondary structure elements through aligned numbers. Loop regions (L) is 1, helices (H) 2 and sheets (C) 4}
\label{fig:sq_mat_struc_sec}
\end{figure}

\subsubsection{Solvent Inaccessibility}

The hydrophobic interactions of non-polar side chains in amino-acids are believed to contribute significantly to the stability of the tertiary structures in proteins. Hydrophobic amino-acids will tend to cluster together, not as a result of attraction, but as a result of their repulsion by the hydrogen bond water network in which the protein is dissolved. Therefore, these amino-acids will preferentially be located away from the surface of the molecule. Since they form the core of protein, they tend to be more conserved and are, thus, more useful for identifying remote evolutionary relationships.

We have utilized the PSA \citep{Lee71} program to provide solvent inaccessibility information. PSA is part of the JOY package. The $M_{s}$ matrix was built giving weight 3 for inaccessible residues and weight one to the others. The weights are based on \citep{Chakrabarti04}, which demonstrated empirically that inaccessible amino-acids are three times more conserved than accessible amino-acids. The $M_{s}$ matrix represents structural weights that were used to build the model \textit{pHMMAcc}, as shown in figure~\ref{fig:hmmer_struct}.
\\

\subsubsection{Packing density}
The tertiary structure of proteins stems from a very large number of atomic interactions. In regions where the interactions are stronger residues tend to be packed together.  It is well known that densely packed regions tend to be preserved, and hence that amino-acids belonging to those regions are usually more conserved than other amino-acids. TJ Ooi created a measure, called the Ooi Number \citep{Ooi86}, that estimates the amino-acid packing density. Essentially, the Ooi number counts for a residue counts the number of neighboring C-$\alpha$ atoms within a radius of 14\AA{} of the given residue’s own C-$\alpha$. Although crude, this measure does give a good impression of which parts of the structure are buried and which are exposed on the surface.

We again use the JOY package to obtain the Ooi number and estimate packing density. Figure \ref{fig:esq_ooinumber} shows a stretch of JOY output, in which the numbers represent the Ooi measure for the Dehaloperoxidase protein in the Globins family (16wc PDB code). We used these numbers to build the structural weight matrix $M_{s}$. The structural weights were than used to build the model \textit{pHMMOoi}, as shown in figure~\ref{fig:hmmer_struct}.

\begin{figure}[!h]
\centering
\includegraphics[scale=0.6]{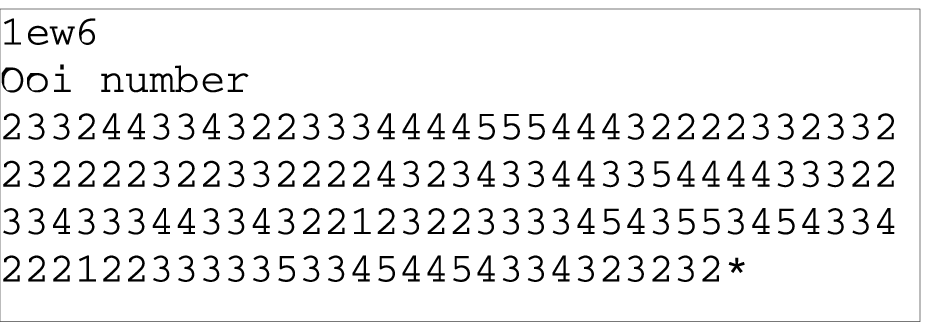}
\caption{Ooi measure for the Dehaloperoxidase protein of Globins family (16wc PDB code), each number represents the amount of neighbor amino-acids inside a radius of 14\AA{}.}
\label{fig:esq_ooinumber}
\end{figure}

\subsubsection{Homologuous Core Structure} 

Structural similarity among proteins can provide valuable insights into their functionality.  One way to provide structural similarities is through three-dimensional alignment of proteins, also called \emph{structural alignment}. The goal is to align two or more proteins by trying to overlap the three-dimensional coordinates of their atoms. When multiple homologue proteins are structurally aligned, we tend to observe that there is a subset of coordinates whose spatial locations are better conserved across structural alignment. This subset is called the \emph{homolog core structure (HCS)} \citep{Matsuo99}. According to the result reported by \cite{HSC_Gerstein95}, HCS can be utilized to detect homologue proteins.

Our goal was to estimate the HCS of a set of protein. As a first approximation, we propose a method to extract it from structural alignment by calculating how much aligned residues from different proteins tend to be close together. Following MAMMOTH, we represent residues through the coordinates of their C-$\alpha$ atoms. In other words, we assume that closeness between C-$\alpha$ atoms will approximate overlapping among amino-acids. To find out how much amino-acids are close together, we utilize the Euclidian distance measure, as shown in the equation \ref{eq:euclid_distance}. It represents the shortest distance between two points in the space.
\\

\begin{equation}
de_{a,b} = \sqrt{(x_{a} - x_{b})^{2} + (y_{a} - y_{b})^{2} + (z_{a} - z_{b})^{2}}
\label{eq:euclid_distance}
\end{equation}
\\
The degree of overlap between aligned residues in the structural alignment was calculated through the relative distance $di_{j}$, equation \ref{eq:relative_distance}. This distance can be found through the average distance among the amino-acid in the position $ij$ and other amino-acids in the $j$ column of alignment.

\begin{equation}
di_{j} =  \dfrac{\sum_{b=j}^{n-1}de_{(i,b),(i,b+1)}}{n-1}
\label{eq:relative_distance}
\end{equation}
\\
Finally, the relative distance was normalized according to \ref{eq:normalization}, and it was used to determine the degree of overlap of each residue. These measures were normalized by using the equation \ref{eq:normalization}, where $d_{min}$ is the minimal distance and $O_{max_{i}}$ is the maximal Ooi measure for protein $i$.
\\
\begin{equation}
m_{ij} =  \dfrac{d_{min}*O_{max_{i}}}{di_{j}}
\label{eq:normalization}
\end{equation}
\\
After this step, we built the $M_{s}$ matrix, where each $m_{ij}$ matrix element corresponds to the relative distance of amino-acids $ij$ in the structural alignment. This matrix represents structural weights that were used to build the model \textit{pHMM3D}, shown in the figure~\ref{fig:hmmer_struct}.
\\

\subsection{Library of structural models}

In a second step, we join the models built from these matrices to form a library of structural models aiming at building a single model to represent the structural patterns under different aspects. We used the \texttt{hmmpfam} HMMER tool to combine the models together. Library of models have been used in a number of studies, such as \citep{PFAM_Eddy04,TIGRFAMsWhite03,Gough01}, and they are known to achieve better results than those achieved by single models.

\subsection{Test Procedure} 
The main concern of our study is to build pHMMs that can be helpful in remote homology detection. Therefore, our experiments considered proteins with identity below 30\%. To do so, we used the SCOP database \citep{SCOP_Murzin04}, and more specifically ASTRAL SCOP version 1.67 PDB40 (with 6600 protein sequences). ASTRAL SCOP is particularly interesting for our study because it describes structural and evolutionary relationships among proteins, such that none of the sequences in ASTRAL SCOP present $>40$\% sequence identity. Thus, it is an excellent dataset to evaluate the performance of remote homology detection methods and has been widely used to reach this goal \citep{STR_Oliva05,HMMERSAMSonnhammer05,RHD_STRBystroff04,Alexandrov04}.

SCOP classifies all protein domains of known structure into a hierarchy with four levels: class, fold, super family and family. In our study, we work at the super family level, which gathers families in such a way that a common evolutionary origin is not obvious from sequence identity, but probable from an analysis of structure and from functional features. We believe that this level better represents remote homolog.

Moreover, we used cross-validation~\citep{Mitchell97} to compare the different approaches. First, we divided SCOP database by super family level. Next, from ASTRAL PDB40, we chose those super families containing at least three families and at least 20 sequences. We eventually tested 39 super families, as listed in Table 1. This whittled down the number of sequences we used to model building to 1137. Third, we implemented leave-one-family-out cross-validation. For any super family $x$ having $n$ families, we built $n$ profiles so that each profile $P$ was built from the sequences in the remaining $n-1$ families. Thus, the $n-1$ sequences form the training set for profile $P$. The test set for profile $P$ will be the remaining sequences (test positives) plus all other database sequences (test negatives).

\begin{table}[!h]

\processtable{Superfamily SCOP-Ids\label{Tab:01}}
{\scriptsize
\begin{tabular}{llllllll}
\toprule
a.1.1.  &   a.138.1. &  a.25.1. &  a.26.1.  &  a.3.1.  & a.39.1.  & a.4.1.   & b.121.4.\\
b.18.1. &   b.29.1.  &  b.36.1. &  b.47.1.  &  b.55.1. & b.60.1.  & b.6.1.   & b.71.1.\\
b.82.1. &   c.1.10.  &  c.23.1. &  c.26.1.  &  c.36.1. & c.52.1.  & c.55.1.  & c.55.3.\\
c.67.1. &   d.108.1. &  d.14.1. &  d.144.1. &  d.15.1. & d.153.1. & d.169.1. & d.3.1.\\
d.58.7. &   d.92.1.  &  g.3.11. &  g.3.6.   &  g.3.7.  & g.37.1.  & g.39.1. & \\\botrule
\end{tabular}}{SCOP Super families used in our experiments. We only considered super families with at least 20 proteins and three or more families.}
\end{table}

In order to assess HMMER-STRUCT performance, we used the HMMER package. We did not compare with SAM \citep{SAM96} package. First, because our goal was to evaluate whether structural properties can improve pHMMs, not to compare the two packages, and second, because a related previous study on the same dataset actually showed HMMER outperforming SAM~\citep{Bernardes07}. The same study also indicated better results on the ``twilight zone'' using structural alignment tools, such as MAMMOTH-mult and 3DCOFFEE. We used MAMMOTH in this study.

Results were graphically analyzed by building ROC and Precision/Recall curves. ROC curves are a common measure of performance that is very used in bioinformatics application. They are based on the relation of the false positives (non homologue proteins) and of true positives (homologue proteins), and are obtained by varying a parameter that affect these relationships.  We further present Precision/Recall curves, as they give a good perspective on true positives, false positives and false negatives hits. In both cases, the bigger the area under the curve (AUC), the more efficient the analyzed tool is. In both cases we used the minimal \emph{e-value} required to accept a match as the parameter used to build both curves. We ranged e-values between $10^{-50}$ and $10$. Finally, we used the paired two tailed t-test to assess significance, and assumed that results with $p \leq 0.05$ (I.e. 95\% of confidence) are significant.

\end{methods}

\section{Results}

\begin{figure}[!h]
\centering
\includegraphics[scale=0.65]{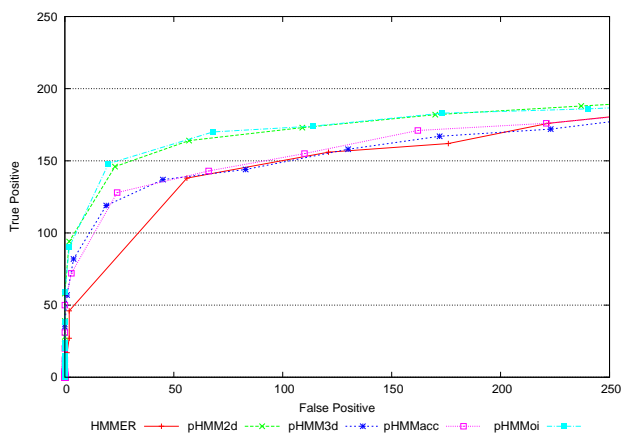}
\caption{Performance of each model in HMMER-STRUCT tool, for MAMMOTH aligner, as measured by ROC Curves}
\label{fig:hmmer_struct_roc_curves}
\end{figure}

\begin{figure}[!h]
\centering
\includegraphics[scale=0.65]{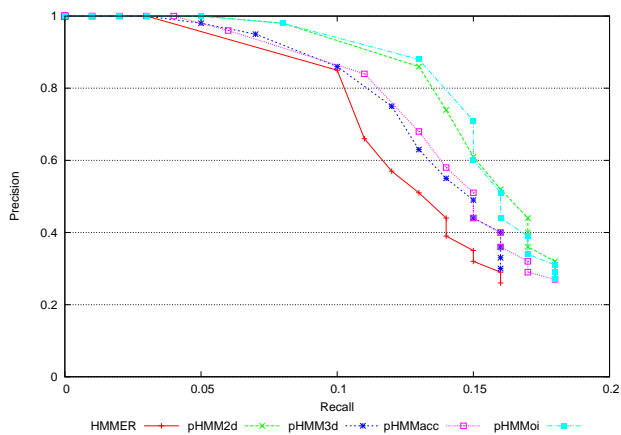}
\caption{Performance of each model in HMMER-STRUCT tool, for MAMMOTH aligner, as measured by Precision/Recall Curves}
\label{fig:hmmer_struct_rp_curves}
\end{figure}

As a first step, we build a model for each structural property and evaluate it according to the methodology described in the Methods section. The ROC curves are presented in figure \ref{fig:hmmer_struct_roc_curves} and the Precision/Recall curves in figure \ref{fig:hmmer_struct_rp_curves}.  Both figures show all models, that is, \textit{pHMM2D} (secondary structural model), \textit{pHMMOi} (Ooi measure model), \textit{pHMMAcc} (inaccessibility model) and \textit{pHMM3D} (three-dimensional structure model) outperforming the HMMER model. Table \ref{Tab:02} shows the paired two tailed t-test between each model. All models built from structural properties perform significantly when compared to HMMER. Only, the \textit{pHMM3D} and \textit{pHMMAcc} results are not significant in relation to each other.


\begin{table}[!h]

\processtable{HMMER-STRUCT paired t-test\label{Tab:02}}
{\scriptsize
\begin{tabular}{|l|l|l|l|l|}
\hline
	&     HMMER  &   pHMM2D   &  pHMM3D     &  pHMMAcc  \\\hline
pHMMOi  &    0,00741 &   0,04     & 0,01  &   0,01    \\\hline
pHMMAcc &    0,03877 &   0,01     & 0,05  &            \\\hline
pHMM3D  &    0,01    &   0,01     &             &             \\\hline
pHMM2D  &    0,00660 &            &             &             \\\hline
\end{tabular}}{Paired two tailed t-test when comparing performance of each HMMER-STRUCT model all against all.}
\end{table}

\begin{figure}[!h]
\centering
\includegraphics[scale=0.7]{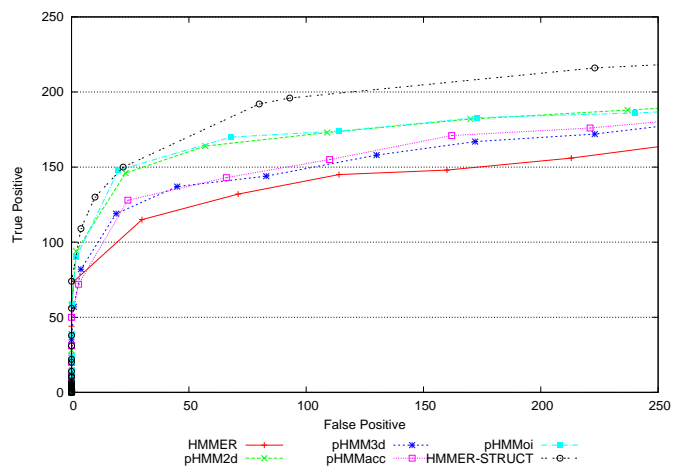}
\caption{HMMER-STRUCT Performance for MAMMOTH aligner, as measured by ROC Curves}
\label{fig:hmmer_struct_pfam_roc_curves}
\end{figure}

\begin{figure}[!h]
\centering
\includegraphics[scale=0.7]{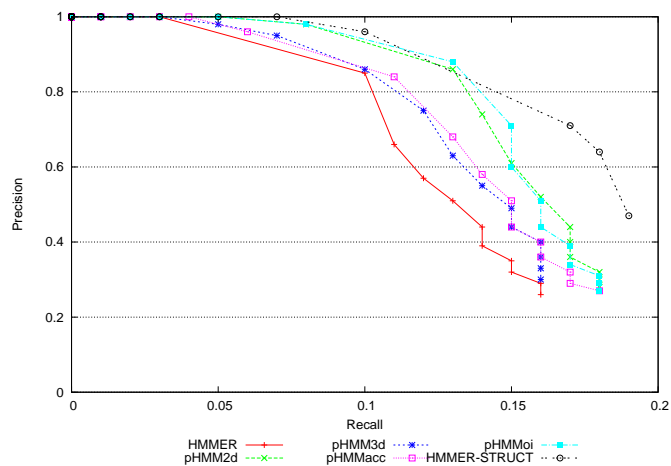}
\caption{HMMER-STRUCT Performance for MAMMOTH aligner, as measured by Precision/Recall Curves}
\label{fig:hmmer_struct_pfam_rp_curves}
\end{figure}

Next, we compare the performance of the model library with respect to the initial HMMER model. To do so, we joined the five models, one for each structural property, and scored the test sequences using \textit{hmmpfam}. Figure \ref{fig:hmmer_struct_pfam_roc_curves} shows the ROC curve for the results. Figure \ref{fig:hmmer_struct_pfam_rp_curves} shows graphically the results through Precison/Recall curves. Both figures show HMMER-STRUCT outperforming HMMER. Table \ref{Tab:03} displays significance results. The difference between HMMER-STRUCT and HMMER results are statistically significant according to paired two tailed t-test. The two tailed t-test also indicate significant differences between HMMER-STRUCT and each HMMER-STRUCT component, i.e, HMMER, \textit{pHMM2D}, \textit{pHMM3D}, \textit{pHMMAcc} and \textit{pHMMOi}.

\begin{table}[!h]

\processtable{HMMER-STRUCT paired t-test\label{Tab:03}}
{\scriptsize
\begin{tabular}{|l|l|}
\hline
		& HMMER-STRUCT \\\hline
HMMER    	& $10^{-4}$	\\\hline
pHMMAcc		& $10^{-4}$	\\\hline
pHMMOi		& $10^{-3}$	\\\hline
pHMM2D		& $10^{-3}$	\\\hline
pHMM3D		& $10^{-4}$	\\\hline
\end{tabular}}{Paired two tailed t-test when comparing performance of each HMMER-STRUCT component with the combined model.}
\end{table}
%
%

\section{Discussion}
The accuracy of homology detection methods is essential for the problem of inferring the function of unknown-function proteins. However, improving accuracy becomes hard when similarity between sequences is low. We proposed a method to improve pHMMs sensitivity by adding structural properties in the model building stage. We showed that the pHMMs trained according to this method are more sensitive than pHMMs trained from multiple sequence alignments, even if the alignment itself relied on structural properties. Our experiments demonstrated best performance for \textit{pHMM2D}, that used secondary structural properties, and for \textit{pHMMOi}, that used packing density residues. Both pHMMs present similar performance. We believe that the good results obtained with the \textit{pHMMoi} model can be attributed to the fact that tight packing is important for protein stability, and follow well-known results that indicate that amino-acids located in the core protein are more conserved than amino-acids located in other sites \citep{Privalov96}. In the same way, the \textit{pHMM2D} model achieve good performance as secondary structure elements are responsible for maintaining the form in homologue proteins. These elements form motifs and domains, which are related with protein function. Conserved sites may point to functionally and structurally important regions. These observations may explain the higher performance of models based on packing residues and on secondary structural properties.

The \textit{pHMMAcc} models, based on amino-acid inaccessibility, and the \textit{pHMM3D} models, based on three-dimensional coordinates, did not perform as well. The \textit{pHMMAcc} models did not achieve statistical significance results, when they were compared with HMMER. On the other hand, we observe that the inaccessibility property can be explained by hydrophobic effects, as are the amino-acids with hydrophobic side-chain that go toward the core protein by forming packages. Therefore hydrophobicity was represented in the \textit{pHMMOi} model, that achieved good performance. Our results suggest the difference between models stems from the \textit{pHMMOi} models to be more accurate and precise than what is used when building \textit{pHMMAcc}.

However, we believe the inaccessibility property is already represented appropriately by \textit{pHMMOi} model. Since amino-acids with high packing density already are inaccessible. Therefore, \textit{pHMMOi} outperformed the \textit{pHMMAcc}, as \textit{pHMMoi} has more information than \textit{pHMMAcc}.

The chief contribution of our method was achieved when all the models work together. The combined models performed significantly better than any single model. We believe that this results from the fact that each trained pHMMs represents a different structural property. Therefore, combining the models increases sensitivity by exploring the different structural properties.

Our method shows that structural information can be added during the training phase of pHMM to improve sensitivity, without much changes to the usage of pHMM methodology, and applied to recently discovered proteins for which there is little structural information.


\section{Conclusion}
The increasing number of studies involving pHMMs and the use of structural information has been quite remarkable \citep{RHD_STRBystroff04,Alexandrov04,HMMSTR00}. Most of these approaches build structural models based on three-dimensional coordinates. In contrast, we present a novel methodology to train pHMMs based on structural alignment and other structural properties using a set of homologue protein sequences. Our method builds five models from an aligned homologue sequence set. Each model represents a different structural property, and the union of the models represent the structural context of aligned proteins. The properties used were primary, secondary and tertiary structures, accessibility and packing residue. Note that previous attempts have already used secondary and tertiary structural properties to train pHMM, though in quite a different way. However, accessibility and packing residue properties were used for the first time in pHMM training, with good results in the latter case.

In order, to build each model, we developed a novel sequence-weighting algorithm based on structural weights that are attributed for each amino-acid. Traditional weighting-algorithm works gives the same weight for every residue in the protein. Instead, we propose a method that gives a different weight to each amino-acid into a protein, according to structural properties that suggest it may be in a conserved region. Our results relied on prior work \citep{Chakrabarti04,Deane03,Ooi86} that suggested interesting properties and estimated their weight.

Nowadays, the most popular approach to discovering the function of a newly found protein is through sequence similarity search. In fact, it is well known that structure is more conserved than sequence, and thus structural similarity can suggest function similarity. On the other hand, structural data is sparse and are usually not available for proteins with unknown function. Therefore, it is very important that methods that uses structural properties to build models will not need to rely on structural information for a new protein. Our method makes use of structural properties only at the model building stage, but not at scoring.

Our results show that the use of structural properties can improve the sensitivity of remote homology methods. Moreover, the combination of different model (one for each property) outperforms the use of individual properties. A number of future research directions present themselves. It will be interesting to include more models, such as that based on bond-hydrogen properties. Also, it will be interesting to apply our methodology to other remote homology tools, such as SAM \citep{SAM96} and T-HMM \cite{THMM04}. Ultimately, we believe that our work is a step in the major challenge of finding the set of structural properties or features that represent precisely membership of a super family.


\section*{Acknowledgement}
We are grateful to CNPq for financial support.

\end{document}